\documentclass{article}
 \PassOptionsToPackage{numbers, compress}{natbib}
 \usepackage[preprint]{neurips_2025}

\usepackage[utf8]{inputenc} 
\usepackage[T1]{fontenc}    
\usepackage{hyperref}       
\usepackage{url}            
\usepackage{booktabs}       
\usepackage{amsfonts}       
\usepackage{nicefrac}       
\usepackage{microtype}      
\usepackage{xcolor}         

\usepackage{makecell}
\usepackage{graphicx}
\usepackage{pifont}
\usepackage{enumitem}
\usepackage{amsmath}
\usepackage{wrapfig}

\title{Align Beyond Prompts: Evaluating World Knowledge Alignment in Text-to-Image Generation}

\author{%
  Wenchao Zhang\textsuperscript{\rm 1}\thanks{Equal Contribution.} \quad
  Jiahe Tian\textsuperscript{\rm 1}$^*$ \quad 
  Runze He\textsuperscript{\rm 1} \quad 
  Jizhong Han\textsuperscript{\rm 1} \quad 
  Jiao Dai\textsuperscript{\rm 1}\\
  \textbf{Miaomiao Feng}\textsuperscript{\rm 1}\quad
  \textbf{Wei Mi}\textsuperscript{\rm 1}\quad
  \textbf{Xiaodan Zhang}\textsuperscript{\rm 1} \\
  \textsuperscript{\rm 1}Institute of Information Engineering, Chinese Academy of Sciences \\
}

\begin{document}

\maketitle
\begin{figure}[ht]
  \centering
\
\includegraphics[width=1\textwidth]{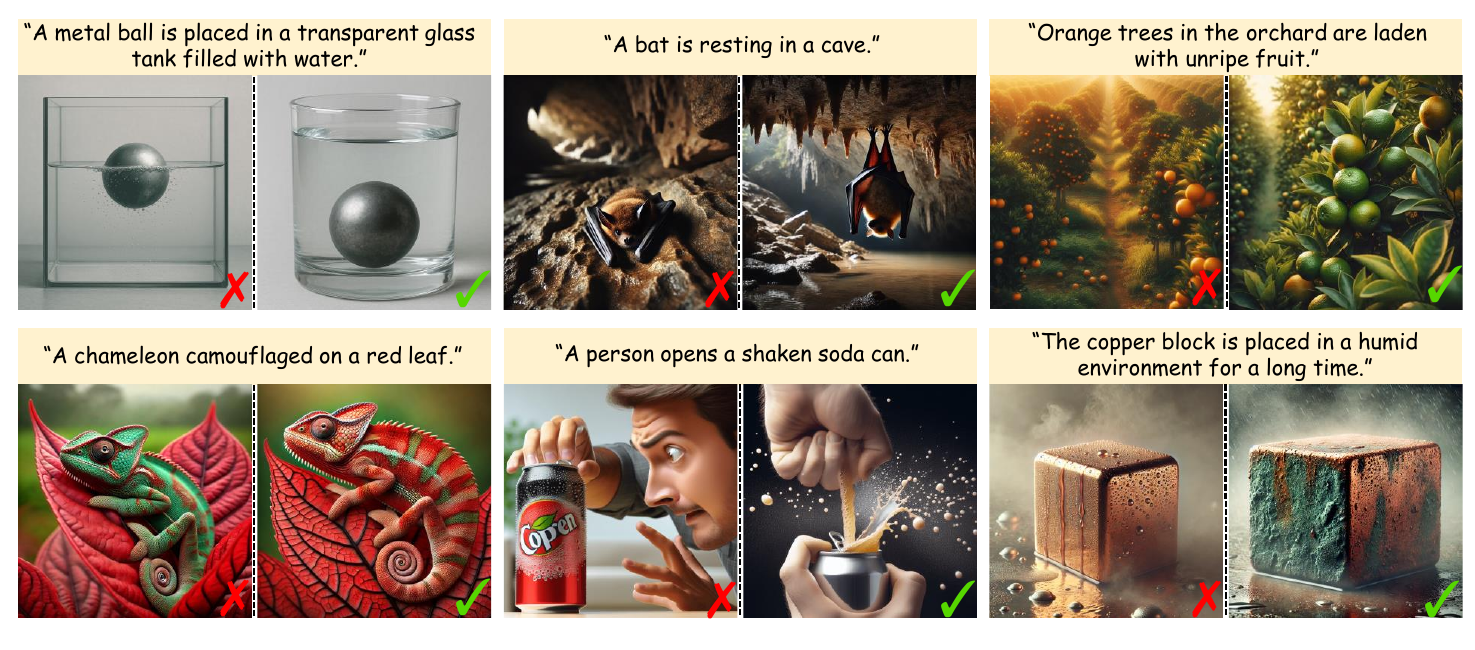}
  \caption{\textbf{Examples of \textsc{ABP}}, example presents images before (fail to align with real-world knowledge) on the left, and after (correction of alignment with world knowledge) optimization using \textsc{ITKI} on the right. Each image includes the \textsc{ABPScore} result, with correctly generated images marked with a check mark "\checkmark", while incorrect images are marked with a cross "\ding{55}".}
  \label{figure:examples}
\end{figure}

\begin{abstract}
  Recent text-to-image (T2I) generation models have advanced significantly, enabling the creation of high-fidelity images from textual prompts. However, existing evaluation benchmarks primarily focus on the explicit alignment between generated images and prompts, neglecting the alignment with real-world knowledge beyond prompts. To address this gap, we introduce \textit{Align Beyond Prompts} (\textsc{ABP}), a comprehensive benchmark designed to measure the alignment of generated images with real-world knowledge that extends beyond the explicit user prompts. \textsc{ABP} comprises over 2,000 meticulously crafted prompts, covering real-world knowledge across six distinct scenarios. We further introduce \textsc{ABPScore}, a metric that utilizes existing Multimodal Large Language Models (MLLMs) to assess the alignment between generated images and world knowledge beyond prompts, which demonstrates strong correlations with human judgments. Through a comprehensive evaluation of 8 popular T2I models using \textsc{ABP}, we find that even state-of-the-art models, such as GPT-4o, face limitations in integrating simple real-world knowledge into generated images. 
  To mitigate this issue, we introduce a training-free strategy within \textsc{ABP}, named \textit{Inference-Time Knowledge Injection} (\textsc{ITKI}). By applying this strategy to optimize 200 challenging samples, we achieved an improvement of approximately 43\% in \textsc{ABPScore}.
  The dataset and code are available in \href{https://github.com/smile365317/ABP}{\texttt{github.com/smile365317/ABP}}.
\end{abstract}

\section{Introduction}
Recent advancements in text-to-image (T2I) generation models \cite{ramesh2021zero,DBLP:conf/cvpr/KimKY22a,esser2024scaling,betker2023improving} have enabled the creation of visually realistic images that align with user-provided textual prompts. 
However, current T2I models still face the challenge of accurately aligning with world knowledge beyond the provided prompts, such as commonsense and factual knowledge.
As illustrated in Figure \ref{figure:examples}, these models often fail to integrate basic world knowledge (e.g., a metal ball should sink to the bottom of water, bats typically rest in an inverted position, etc.) into the generated visual content. 
However, recent work has focused on designing benchmarks and metrics to evaluate specific capabilities of T2I models, such as aesthetic quality \cite{saharia2022photorealistic, schuhmann2022laion}, implausibility \cite{liang2024rich, wu2023human, xu2023imagereward, kirstain2023pick}, image-text alignment \cite{li2024genai, hu2023tifa, cho2023davidsonian, lin2024evaluating,wiles2024revisiting}, compositionality \cite{huang2023t2i, wu2024conceptmix,zhu2023contrastive,yu2022scaling}, numerical reasoning \cite{kajic2024evaluating,paiss2023teaching,acharya2019tallyqa}, and spatial reasoning \cite{gokhale2022benchmarking}.
Nevertheless, there is no comprehensive benchmark for evaluating the alignment between generated images and world knowledge beyond the textual prompts.
Generating visually realistic yet factually or physically flawed images poses a risk of misinformation and undermines trust, thereby limiting the safe use of T2I models in applications where real-world accuracy is critical.

\textbf{Challenges. }Developing a comprehensive and rigorous T2I benchmark for evaluating the alignment between generated images and world knowledge beyond the textual prompts presents several challenges.
Firstly, constructing effective prompts for evaluating this task is inherently complex. 
The prompts must implicitly incorporate visually perceivable world knowledge.
For instance, consider the prompt \textit{placing a metal ball in water}. The expected visual output should be \textit{the metal ball should sink to the bottom of the water}, which is implied as world knowledge rather than explicitly stated.
Constructing such prompts using automated tools, such as GPT-4o, proves challenging. These tools often state the desired outcome directly rather than implicitly embedding the necessary world knowledge. Additionally, these tools may fail to cover the full spectrum of real-world knowledge in repeated use.
Secondly, the vast and diverse nature of real-world knowledge makes it challenging to conduct a comprehensive benchmark. 
While recent works \cite{meng2024phybench,fu2024commonsense,li2025science,lee2023holistic} have attempted to evaluate the alignment between images and commonsense knowledge, their scope is limited to specific categories of knowledge. 
Finally, existing evaluation methods \cite{hessel2021clipscore,zhai2023sigmoid,xu2023imagereward,kirstain2023pick,lin2024evaluating,li2025science} fall short in providing precise metrics for this task. 
Textual prompts convey only limited information, and while the generated images may align with the prompts, they often include additional world knowledge that was not part of the prompt. 
This discrepancy renders existing metrics, which primarily focus on prompt-image alignment, insufficient to evaluate the alignment between generated images and world knowledge beyond the textual prompts.

In this work, we propose \textsc{ABP}, a comprehensive benchmark designed to evaluate the consistency between generated images and world knowledge that extends beyond the explicit user prompts.
Firstly, we developed a systematic prompt creation pipeline to construct prompts that meet the evaluation requirements.
This pipeline begins by collecting visually perceivable knowledge anchors, such as \textit{chameleons can change their color.} These knowledge anchors are then implicitly integrated into specific scene prompts, such as \textit{a chameleon camouflaged on a red leaf.} 
Through this process, we curated 2,060 meticulously crafted prompts that satisfy the necessary evaluation requirements.
Secondly, the collected prompts span six major knowledge domains, ensuring a broad coverage of world knowledge. 
Compared to existing benchmarks \cite{meng2024phybench,fu2024commonsense,li2025science}, \textsc{ABP} offers a more extensive range of world knowledge.
Utilizing \textsc{ABP}, we evaluated eight state-of-the-art T2I models, including GPT-4o \cite{openai2025addendum}, Gemini-2.0-flash-exp-image-generation (Gemini 2.0) \cite{team2023gemini}, DALL-E 3 \cite{betker2023improving}, Midjourney V6 \cite{midjourney2024}, stable-diffusion-3-medium (SD3-M) \cite{esser2024scaling}, stable-diffusion-3.5-large (SD3.5-L) \cite{esser2024scaling}, stable-diffusion-xl-base-1.0 (SDXL) \cite{podell2023sdxl} and CogView4-6B (CogView4) \cite{ding2021cogview}. A total of 22,660 images were generated (four images produced per prompt by Midjourney V6), and 30,867 human judgments were collected.
Finally, to more accurately assess the task, we propose \textsc{ABPScore}. This metric that utilizes existing Multimodal Large Language Models (MLLMs) to measure the alignment between the generated images and world knowledge beyond the textual prompts.
Specifically, \textsc{ABPScore} leverages commonsense knowledge to infer key behaviors in the generated visual content and utilizes MLLMs to verify these behaviors. This metric demonstrates superior correlations with human judgments compared to existing metrics. Our results indicate that current T2I models face significant challenges in generating world knowledge beyond the textual prompts. 
We additionally propose a training-free strategy within \textsc{ABP}, referred to as \textit{Inference-Time Knowledge Injection} (\textsc{ITKI}), which yields a substantial enhancement in performance. By applying this strategy to optimize 200 challenging samples, we achieved an approximately 43\% improvement in \textsc{ABPScore}, significantly mitigating the issue.

To sum up, our main contributions can be summarized as follows:
\setlength{\leftmargini}{0.85em}
\begin{itemize}[topsep=0pt, itemsep=0pt]
\item We introduce the \textsc{ABP} benchmark, comprising 2,060 meticulously curated prompts, and collect 30,867 human judgments to systematically assess the alignment of generated images with real-world knowledge beyond the prompts.
\item We propose \textsc{ABPScore}, a metric that leverages existing MLLMs for evaluation, demonstrating superior correlations with human judgments compared to existing metrics.
\item We further introduce \textsc{ITKI}, a model-agnostic strategy within \textsc{ABP} to effectively comprehends world knowledge in generated images, which demonstrates effectiveness across existing benchmarks.
Especially, experiments conducted on 200 most challenging samples from the \textsc{ABP} demonstrate that our strategy yields an approximate 43\% improvement in \textsc{ABPScore}.
\end{itemize}

\section{Related works}

\textbf{Text-to-Image (T2I) generation models.}
T2I generation models are generally trained to generate images based on textual prompts. 
Starting from DALL-E \cite{betker2023improving}, T2I generation models began to demonstrate impressive text prompt following capabilities, with widely used GANs \cite{goodfellow2020generative} as the visual generation module. 
Subsequently, diffusion models \cite{rombach2022high,podell2023sdxl,esser2024scaling} have achieved remarkable success in T2I models. 
Early diffusion-based T2I methods typically injected text conditions into the UNet or DiT networks via cross-attention or AdaLN mechanisms. 
More recently, several works \cite{openai2025addendum,team2023gemini} directly integrated multi-modal large language models (MLLMs) with T2I models, allowing for more flexible inputs, such as long text and multiple reference images.
Besides, some works \cite{sanjay2024enhancing,cohen2022diffusion} have also explored using autoregressive models and VQ decoding to directly generate images without employing the diffusion process.
As the image generation module continues to advance, the text prompt encoders used in T2I generation models have also improved, progressing from early vision-language models such as CLIP \cite{esser2024scaling} and BLIP \cite{li2023blip} to contemporary LLMs \cite{openai2025addendum,team2023gemini}.
These improvements significantly enhancing the text prompt following and commonsense reasoning capabilities of T2I generation.
However, we observed that even the most advanced text-to-image models can still generate images that fail to accurately infer the expected visual outcomes in the generated images that merely requires simple commonsense reasoning.

\textbf{Benchmarking text-to-image (T2I) models.}
Previous evaluations of text-to-image (T2I) generation models have primarily focused on the visual quality of generated images and the alignment between generated images and text prompts. 
Traditional assessment of visual quality emphasized fidelity metrics such as Fréchet Inception Distance (FID) \cite{heusel2017gans} and Inception Score (IS) \cite{salimans2016improved}, and subsequent work includes additional attributes such as aesthetic quality \cite{schuhmann2022laion} and plausibility \cite{liang2024rich}. 
For text-image alignment, early approaches \cite{kumari2023multi,ruiz2023dreambooth,wu2023tune,bugliarello2023measuring} typically employed vision-language similarity metrics, with CLIPScore \cite{hessel2021clipscore} being a representative example.
Some studies \cite{wu2023human,xu2023imagereward,kirstain2023pick,liang2024rich,yarom2023you} have also investigated the use of reward models trained on human preference data to evaluate image-text consistency.
Recently, several studies \cite{ku2023viescore,zhang2023gpt} have begun utilizing MLLMs to directly assess image-text consistency. 
Alternative approaches \cite{hu2023tifa,cho2023davidsonian,li2024evaluating,lu2023llmscore} have developed visual question answering (VQA) pipelines based on MLLMs to assess image-text consistency.
Recently, few benchmarks have emerged \cite{meng2024phybench,fu2024commonsense,li2025science} focusing on commonsense knowledge and scientific knowledge, yet their evaluation scope remains limited.
Current benchmarks fail to adequately measure the alignment of generated images with real-world knowledge that extends beyond the prompts.
To address this gap, we introduce \textsc{ABP}, a comprehensive benchmark specifically designed to evaluate the alignment of generated images with real-world knowledge beyond the scope of the prompts.

\begin{figure}
  \centering
\includegraphics[width=1\textwidth]{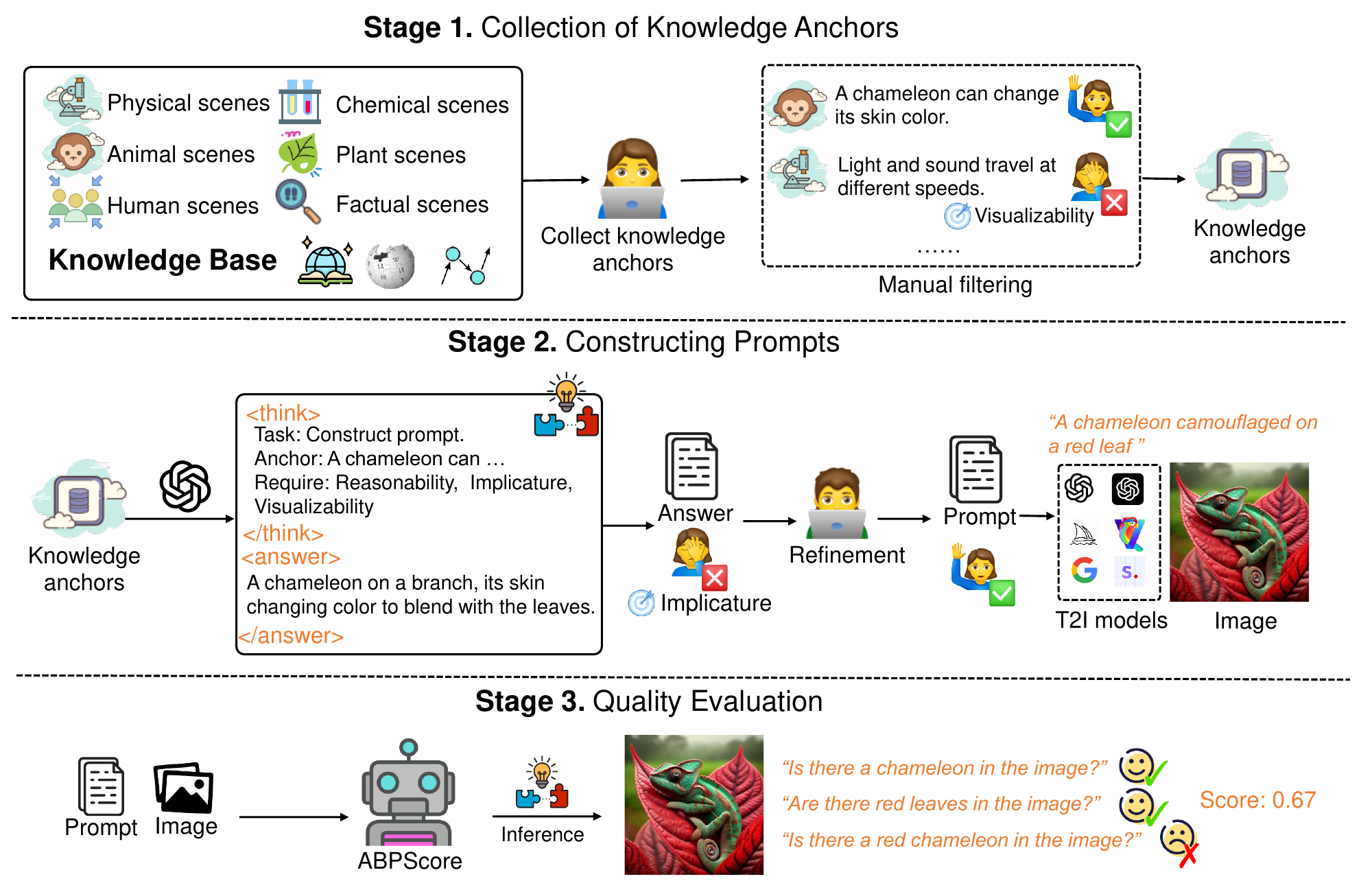}
  \caption{\textbf{The construction of \textsc{ABP}.} 
  (Upper) \textbf{Collection of Knowledge Anchors.} We manually collect and filter knowledge anchors from various online repositories, including Wikipedia and ConceptNet, across six different scenes;
  (Middle) \textbf{Constructing Prompts.} We use GPT-4o to generate prompts from the collected knowledge anchors, which are then filtered and optimized to align with the criteria of \textit{Reasonability}, \textit{Implicature}, and \textit{Visualizability}. Subsequently, images are generated using eight state-of-the-art T2I models;
  (Bottom) \textbf{Quality Evaluation.} We use \textsc{ABPScore} to extract world knowledge beyond the prompts and associated objects, and validate the alignment between the extracted knowledge and the generated images.
  }
  \label{figure: pipeline}
\end{figure}
\section{\textsc{ABP}}

\subsection{Problem Definition} \label{sec:3.1}
In the \textsc{ABP} benchmark, our primary objective is to assess the alignment between the images generated by T2I models and world knowledge that extends beyond the provided prompts.
Overall, the data in \textsc{ABP} can be formulated as a set of triplets, $\langle I, T, R \rangle$, where $T$ denotes the textual prompt, which implicitly incorporates world knowledge $R$, and $I$ represents the image generated based on the prompt. 
The prompt $T$ must satisfy three fundamental criteria: \textbf{Reasonability}, \textbf{Implicature}, and \textbf{Visualizability}. Reasonability ensures the prompt is consistent with established world knowledge, avoiding contradictions with facts or commonsense. Implicature requires that the prompt implicitly incorporates world knowledge, such as commonsense and factual information. Visualizability requires that the world knowledge beyond the prompt aligns with visual content perceptible to humans.

Utilizing automated tools, such as GPT-4o, often fails to generate prompts simultaneously satisfying the three aforementioned criteria. 
Therefore, constructing prompts within \textsc{ABP} that meet these requirements necessitates both specialized domain expertise and meticulous manual validation, making the prompt construction process for \textsc{ABP} particularly challenging.

\subsection{\textsc{ABP} Construction Pipeline} \label{sec:3.2}
In the process of prompt construction for \textsc{ABP}, we focus on six distinct categories of world knowledge, including physical scenes, chemical scenes, animal scenes, plant scenes, human scenes, and factual scenes.
Constructing prompts that simultaneously meet the criteria of \textbf{Reasonability}, \textbf{Implicature}, and \textbf{Visualizability} is a complex task. To address these challenges, we employ a systematic, step-by-step approach.
First, we identify and extract knowledge anchors that align with the characteristics of Reasonability and Visualizability. 
Then, these knowledge anchors are seamlessly integrated into specific scene prompts, while ensuring that the Implicature of the knowledge is preserved. 
Below, we detail the construction of \textsc{ABP}, with an overview of our pipeline provided in Figure \ref{figure: pipeline}.

\textbf{Stage 1: Collection of Knowledge Anchors.}
One of the primary objectives of \textsc{ABP} is to evaluate whether generated images conform to various world knowledge.
To broaden the scope of the evaluation, we enlisted six human experts to systematically gather a substantial number of knowledge anchors containing commonsense or factual knowledge from various online knowledge repositories, including Wikipedia, ConceptNet \cite{speer2017conceptnet}, across the six defined scenes.
These knowledge anchors were then subjected to a manual filtering process, wherein only those that met the criteria of Reasonability and Visualizability were retained. 
For instance, the knowledge anchor \textit{A chameleon can change its skin color} was retained, as it aligns with both criteria.
However, \textit{Light and sound travel at different speeds} was excluded for failing to meet the Visualizability criterion.
Through this rigorous collection and filtering process, we ultimately compile a set of approximately 4,000 knowledge anchors.
Further details on the knowledge anchors are provided in the Appendix.

\begin{table}
  \caption{\textbf{Comparison of different benchmarks.}
  Compared to other benchmarks, \textsc{ABP} covers a broader range of world knowledge scenes. 
  While Science-T2I \cite{li2025science} includes the largest number of prompts, it only encompasses 16 specific tasks, limiting the diversity of world knowledge.
  Both \textsc{ABP} and PhyBench \cite{meng2024phybench} provide human annotations in their datasets to benchmark evaluation metrics.
  }
  \label{overview-table}
  \centering
  \resizebox{\textwidth}{!}{
  \resizebox*{0.9\linewidth}{!}
{
    \begin{tabular}{c c c c c c c c c}
      \hline
      \textbf{Benchmarks} & \textbf{\makecell{Physical\\Scenes}} & \textbf{\makecell{Chemical\\Scenes}} & \textbf{\makecell{Animal\\Scenes}} & \textbf{\makecell{Plant\\Scenes}} & \textbf{\makecell{Human\\Scenes}} & \textbf{\makecell{Factual\\Scenes}} & \textbf{Number} & \textbf{\makecell{Human\\Annotations}} \\
      \hline
      PhyBench \cite{meng2024phybench} & \checkmark & \ding{55} & \ding{55} & \ding{55} & \ding{55} & \ding{55} & 700   & \checkmark \\
      Commonsense-T2I \cite{fu2024commonsense} & \checkmark & \ding{55} & \checkmark & \checkmark & \checkmark & \ding{55} & 150   & \ding{55} \\
      Science-T2I \cite{li2025science} & \checkmark & \checkmark & \checkmark & \checkmark & \ding{55} & \ding{55} & \textbf{9,000} & \ding{55} \\
      \hline
      \textbf{\textsc{ABP} (Ours)} & \checkmark & \checkmark & \checkmark & \checkmark & \checkmark & \checkmark & 2,060 & \checkmark \\
      \hline
    \end{tabular}}
  }
\end{table}

\textbf{Stage 2: Constructing Prompts.}
After obtaining the knowledge anchors, we integrate them into prompts while preserving the Implicature. To alleviate the manual workload, we automate this process using GPT-4o. Specifically, we provide GPT-4o with a task, the knowledge anchors, the specific requirements (Reasonability, Implicature, Visualizability), and several examples to generate the desired prompts.
We observed that GPT-4o encounters challenges when generating prompts with the Implicature feature. For instance, in the response \textit{A chameleon on a branch, its skin changing color to blend with the leaves}, the knowledge anchor is explicitly stated.
We filter out prompts that do not meet the Implicature criteria and refine them to generate prompts that satisfy all three characteristics.
Through filtering and refinement processes, 2,060 specific prompts were generated, with each prompt incorporating multiple knowledge anchors to enhance its complexity. 
Using the aforementioned eight state-of-the-art T2I models, 22,660 images were produced (with Midjourney V6 generating four images for each prompt).

\textbf{Stage 3: Quality Evaluation.}
We introduce an automated metric, \textsc{ABPScore}, which utilizes GPT-4o to assess the alignment between generated images and world knowledge beyond the user-provided prompts.
\textsc{ABPScore} comprises two primary components: extracting and validating world knowledge.
During the extraction process, world knowledge \(\{R_1, R_2, R_3, \dots\}_{i=1}^{N}\) is extracted from the prompt, with the average value of $N$ being approximately 8.9.
We consider world knowledge beyond the prompt and the knowledge of the associated entity.
For instance, the prompt \textit{A chameleon camouflaged on a red leaf} implies the commonsense knowledge that \textit{the chameleon's skin is red}. However, the prerequisite conditions for this knowledge are the \textit{existence of a chameleon} and the \textit{presence of a red leaf}.
Finally, we validate whether the generated image accurately represents the extracted world knowledge. The \textsc{ABPScore} is defined as:
\[
\textsc{ABPScore} = \frac{1}{N} \sum_{i=1}^{N} \mathbb{I} \left( MLLM(I, R_i) = A_i \right),\tag{1}
\]
where \( I \) represents the generated image, \( R_i \) denotes the world knowledge extracted from the prompt, and \( \mathbb{I}(\cdot) \) is an indicator function that returns 1 if the condition is satisfied and 0 otherwise. \( MLLM \) is a model capable of verifying whether the world knowledge \( R_i \) is present within the image \( I \), and \( A_i \) represents the ground truth. The value of \textsc{ABPScore} is in the range \([0, 1]\).

\subsection{Characteristics and Statistics.}
\begin{figure}
  \centering
\includegraphics[width=1\textwidth]{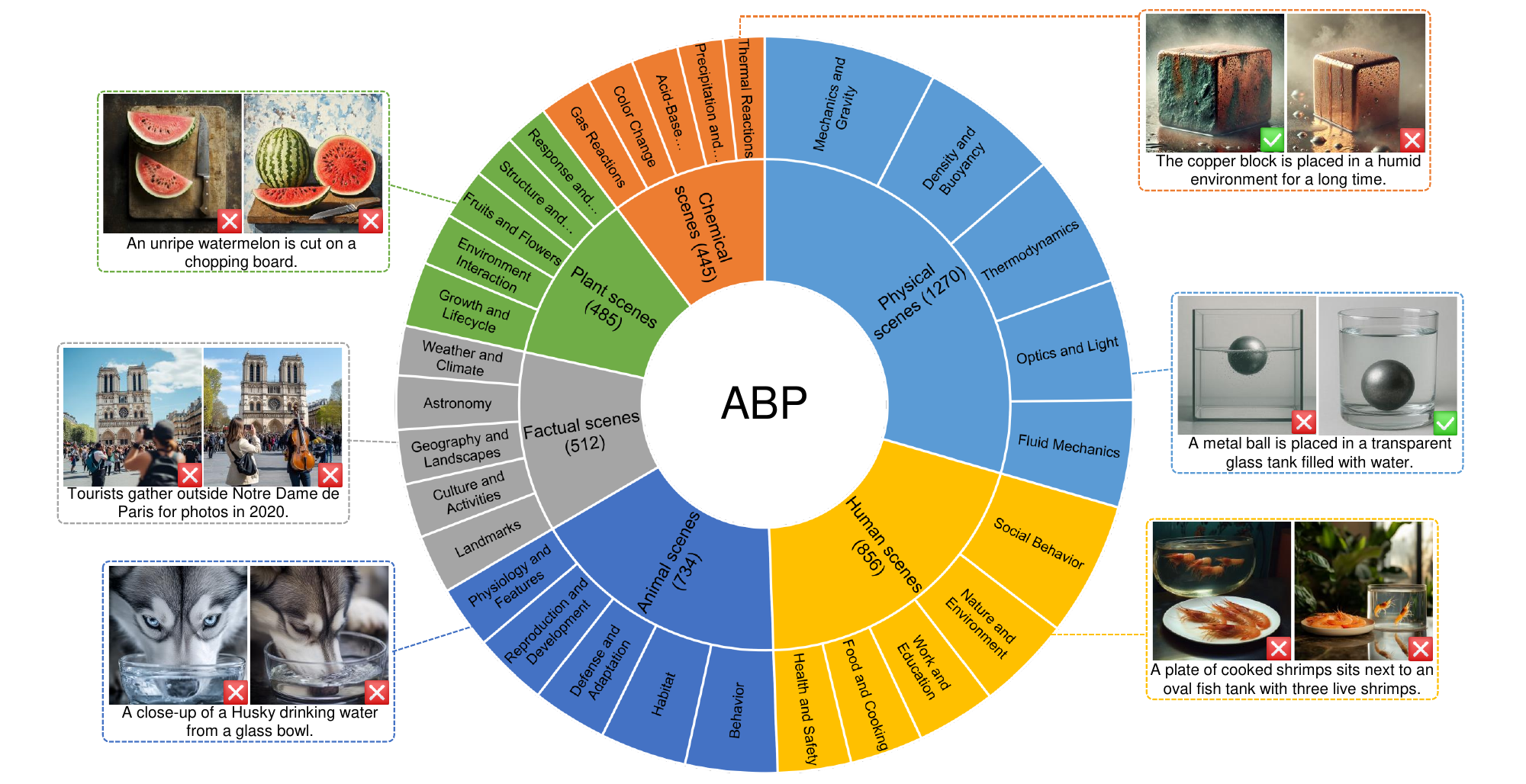}
  \caption{\textbf{Statistics for the \textsc{ABP} dataset.} 
  The inner ring illustrates the six world knowledge domains covered by \textsc{ABP}: physical scenes, chemical scenes, animal scenes, plant scenes, human scenes, and factual scenes.
  As individual prompts may span multiple knowledge domains, the total number of prompts across all domains exceeds 2,060.
  The outer ring illustrates the five most frequent specific knowledge categories within each domain.
  }
  \label{figure: dataset}
\end{figure}

\textsc{ABP} consists of 2,060 carefully crafted prompts and 22,660 generated images, covering six distinct scenes, with each prompt incorporating implicit world knowledge.
To the best of our knowledge, this is the most comprehensive publicly available benchmark for evaluating world knowledge beyond the prompts (a comparison with other benchmarks is provided in Table \ref{overview-table}). The distribution of prompt quantities and the top five knowledge anchors for each scene are illustrated in Figure \ref{figure: dataset}.

\subsection{Human Judgments via \textsc{ABP}} \label{sec:3.4}
\textbf{Human judgments.}
We first utilize eight T2I models—GPT-4o, DALL-E 3, Gemini 2.0, Midjourney V6, CogView4, SD3.5-L, SD3-M, and SDXL—to generate 22,660 images.
Next, we hire three evaluators to manually annotate the extent to which these images conform to world knowledge.
The judgment methodology employs a 5-point Likert scale \cite{otani2023toward}, where a score of 1 denotes \textit{does not match at all,} 2 denotes \textit{significant discrepancies,} 3 denotes \textit{several minor discrepancies,} 4 denotes \textit{a few minor discrepancies,} and 5 denotes \textit{matches exactly.} 
Before the judgment process, we trained the evaluators to ensure consistency in the judgment criteria.
Detailed judgment guidelines can be found in the Appendix.

\textbf{Filtering.} 
We observe that differences in evaluators' ability to recognize implicit world knowledge result in slight score variations for specific image-text pairs. This variability is inherent and unavoidable.
According to our judgment criteria, a score difference greater than 2 indicates a substantial divergence in the evaluators' understanding of implicit world knowledge, and such scores are therefore excluded.
Following this filtering process, 30,867 human judgments are obtained. The human judgments we collected demonstrate a high degree of inter-evaluator agreement, with Krippendorff's Alpha value reaching 0.75, indicating substantial consistency among evaluators \cite{hu2023tifa}.

\begin{figure}
  \centering
\includegraphics[width=1\textwidth]{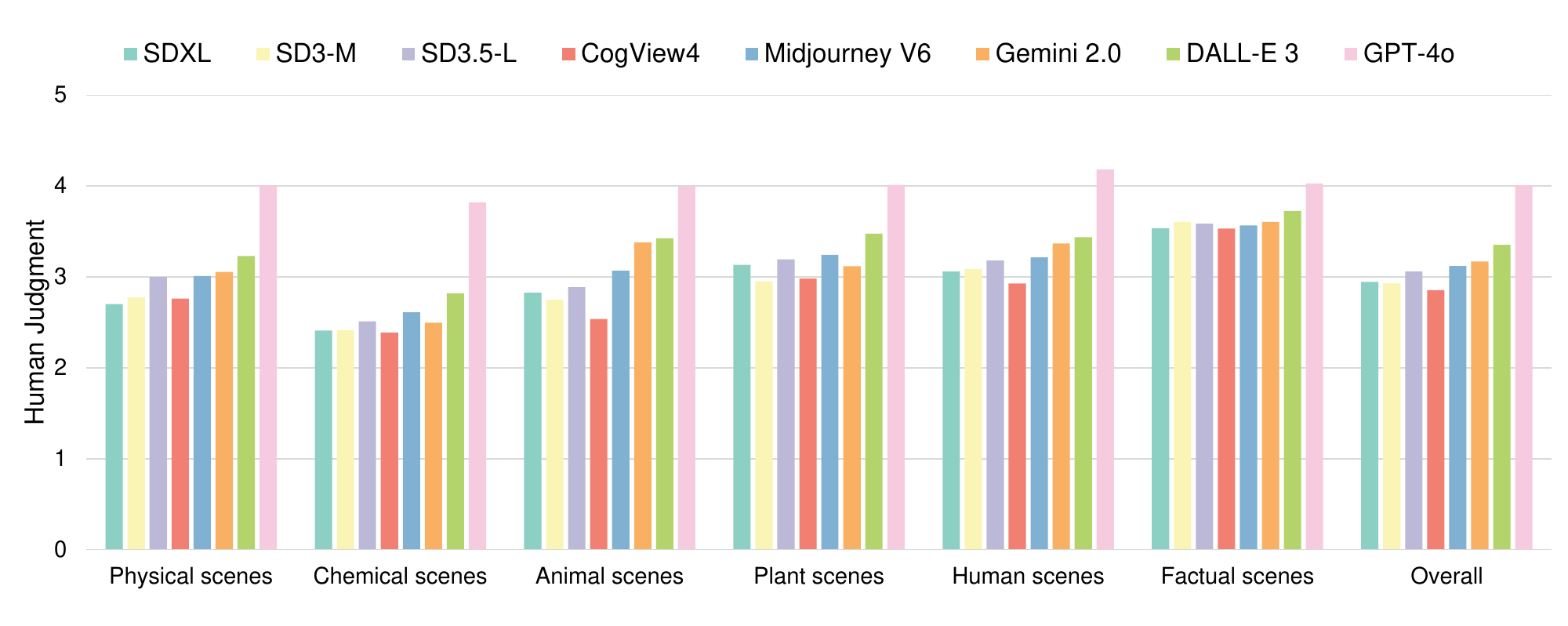}
  \caption{\textbf{Human judgments.}
  We show the average human judgments for eight T2I models, with the first four being open-source (SDXL, SD3-M, SD3.5-L, CogView4) and the remaining four being closed-source (Midjourney V6, Gemini 2.0, DALL-E 3, GPT-4o). 
  Our analysis reveals two key insights: (1) all models demonstrate strong performance in factual scenes, but their performance is significantly weaker in chemical scenes, (2) open-source models still lag behind closed-source models.
  }
  \label{figure:human}
\end{figure}

\textbf{Analysis.}
Figure \ref{figure:human} presents human judgments on generated images using the prompts in \textsc{ABP}.
Among the models evaluated, GPT-4o achieved the highest average score of 4.01, indicating "a few minor discrepancies" between images generated by GPT-4o and world knowledge.
All models demonstrated strong performance in factual scenes, yet their performance in chemical scenes was notably weaker. 
Furthermore, we observed that the performance of the assessed open-source models is directly correlated with the capabilities of the text encoders they employ.

\section{Experiments}
\subsection{Experimental Setup} \label{sec:4.1}
\textbf{T2I Models.}
In this study, we evaluate eight state-of-the-art T2I models: DALL-E 3 \cite{betker2023improving}, Midjourney V6 \cite{midjourney2024}, CogView4 \cite{ding2021cogview}, SD3.5-L \cite{esser2024scaling}, SD3-M \cite{esser2024scaling}, SDXL \cite{podell2023sdxl}, GPT-4o \cite{openai2025addendum}, and Gemini 2.0 \cite{team2023gemini}. All experiments are conducted on an NVIDIA RTX A6000 GPU. 

\textbf{Evaluation Metrics and Baselines.} 
We introduce \textsc{ABPScore} and human judgments to evaluate the performance of the T2I models in generating images across six distinct scenarios.
As baselines, we use the following metrics: CLIPScore \cite{hessel2021clipscore} and SigLIP \cite{zhai2023sigmoid}, which leverage image and text embeddings to calculate the alignment between generated images and world knowledge. Additionally, we utilize metrics trained on human preference data, including HPS V2 \cite{wu2023human}, ImageReward \cite{xu2023imagereward}, and PickScore \cite{kirstain2023pick}. Furthermore, we incorporate SCISCORE \cite{li2025science}, a contemporaneous work, which directly evaluates whether the generated images meet world knowledge. 

\subsection{Correlation with Human Judgments} \label{sec:4.2}
\begin{wraptable}{r}{0.6\textwidth}
\centering
\setlength{\tabcolsep}{4pt} 
\caption{\textbf{Correlations between each evaluation metric and human judgment on \textsc{ABP}.} We report Spearman's $\rho$ and Kendall's $\tau$, with higher scores indicating better performance for all. The proposed \textsc{ABPScore} demonstrates higher correlation with human judgment than prior metrics.}
\label{tab:comparison}
\begin{tabular}{@{}lc@{\hspace{8pt}}c@{}}
\toprule
\textbf{Method} & \textbf{Spearman's $\rho$} & \textbf{Kendall's $\tau$} \\ 
\midrule
CLIPScore \cite{hessel2021clipscore} & 11.2 & 7.5 \\ 
SigLIP \cite{zhai2023sigmoid} & 16.6 & 10.9 \\ 
HPS V2 \cite{wu2023human} & 10.6 & 7.1 \\ 
ImageReward \cite{xu2023imagereward} & 17.0 & 10.9 \\ 
PickScore \cite{kirstain2023pick} & 19.1 & 12.9 \\ 
SCISCORE \cite{li2025science} & 16.1 & 11.1 \\ 
\textbf{\textsc{ABPScore} (Ours)} & \textbf{43.4} & \textbf{32.3} \\ 
\bottomrule
\end{tabular}
\end{wraptable}

We utilize the Pearson and Kendall correlation coefficients to quantify the correlation between the evaluation metrics and human judgments.
The correlations for each evaluation metric and human judgments are presented in Table \ref{tab:comparison}.
Among all the evaluation metrics, CLIPScore, which is based on image and text embedding extracted by CLIP, shows the lowest correlation with human judgments. 
In contrast, SigLIP, an optimization of CLIPScore, shows an improved correlation with human judgments.
Evaluation metrics based on human preferences, such as HPS V2, ImageReward, and PickScore, show a stronger correlation with human judgments, as they implicitly encode the cognitive experiences of annotators within the training data.
In contrast, the SCISCORE metric, which was developed contemporaneously with our work, shows lower correlation with human judgments. This is primarily due to SCISCORE being trained on only 16 specific tasks, which results in diminished accuracy when the knowledge in the prompts exceeds this scope.
Compared to existing metrics, the proposed \textsc{ABPScore} demonstrates a higher correlation with human judgments, validating the reliability of our proposed metric.

\begin{table}[ht]
\caption{\textbf{Different T2I models' results on \textsc{ABP}.} The score of the highest-performing model is highlighted in bold.}
\label{tab:model_performance}
\centering

\resizebox*{0.9\linewidth}{!}
{
\begin{tabular}{l c c c c c c c}
\hline
\textbf{Models} & \textbf{\makecell{Physical\\Scenes}} & \textbf{\makecell{Chemical\\Scenes}} & \textbf{\makecell{Animal\\Scenes}} & \textbf{\makecell{Plant\\Scenes}} & \textbf{\makecell{Human\\Scenes}} & \textbf{\makecell{Factual\\Scenes}} & \textbf{Overall} \\
\hline
SDXL & 0.6511 & 0.5283 & 0.6282 & 0.6924 & 0.6857 & 0.7489 & 0.6558 \\
SD3-M & 0.7011 & 0.5647 & 0.6257 & 0.6923 & 0.7073 & 0.7528 & 0.6740 \\
SD3.5-L & 0.7091 & 0.5734 & 0.6656 & 0.7259 & 0.7226 & 0.7787 & 0.6959 \\
CogView4 & 0.7205 & 0.6228 & 0.6215 & 0.7132 & 0.7201 & 0.8039 & 0.7003 \\
Midjourney V6 & 0.7153 & 0.5843 & 0.7219 & 0.7553 & 0.7360 & 0.8123 & 0.7208 \\
Gemini 2.0 & 0.7397 & 0.6626 & 0.7129 & 0.7371 & 0.7528 & 0.7753 & 0.7301 \\
DALL-E 3 & 0.7630 & 0.7107 & 0.7738 & 0.8077 & 0.7463 & 0.8346 & 0.7727 \\
GPT-4o & \textbf{0.8180} & \textbf{0.7702} & \textbf{0.8243} & \textbf{0.8421} & \textbf{0.8152} & \textbf{0.8581} & \textbf{0.8213} \\
\hline
\end{tabular}}
\end{table}

\subsection{Benchmarking Text-to-Image Models} \label{sec:4.3}
We assessed the ability of eight state-of-the-art T2I models to generate world knowledge beyond the prompts in six knowledge-intensive scenes using the proposed \textsc{ABPScore}.
The results of the experiment are provided in the Table \ref{tab:model_performance}. Based on these results, we have the following observations:
(1) There are significant differences in the performance of various T2I models across different scenes.
GPT-4o demonstrates the highest performance in all scenarios, both in individual scene evaluations and overall scores. This superior performance suggests that GPT-4o excels in understanding and generating world knowledge beyond the prompts.
DALL-E 3 also shows strong generative ability across all scenes, securing the second position in overall scoring.
Following it are Gemini 2.0 and Midjourney V6.
In contrast, open-source models (CogView4, SD3.5-L, SD3-M, SDXL) display lower scores across various scenes, underscoring the difficulties in generating world knowledge that extends beyond the provided prompts.
(2) The performance of T2I models varies across different scenes. Notably, the factual scenes yield the best results, owing to the frequent occurrence of historical architecture and traditional attire in the training data. In contrast, the chemical scenes show consistently lower performance, reflecting the challenges faced by generative models in accurately understanding and representing chemical knowledge.

\textbf{Analysis.}
Among all T2I models, GPT-4o demonstrates the most exceptional performance, owing to its robust reasoning capabilities.
Reasoning capabilities enable the model to more accurately understand world knowledge beyond the prompts. 
This characteristic offers significant insight for future optimization of T2I models, highlighting the crucial role of enhancing reasoning abilities to mitigate gaps in understanding world knowledge during image generation.

\subsection{Inference-Time Knowledge Injection for Enhancing World Knowledge beyond the Prompts} \label{sec:4.4}
\begin{wrapfigure}{r}{0.6\textwidth} 
    \centering
    \includegraphics[width=0.5\textwidth]{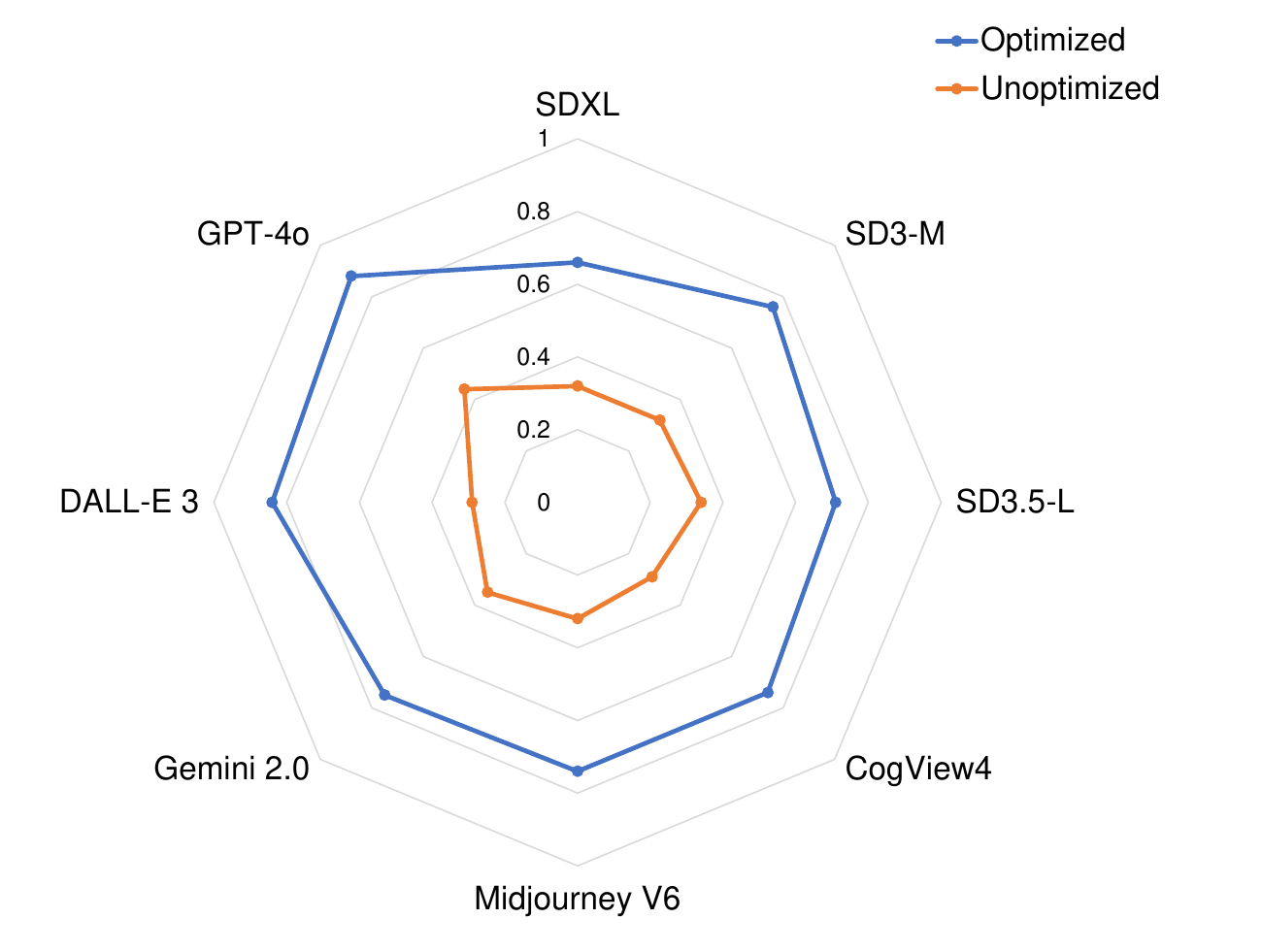}
    \caption{\textbf{Performance Comparison Before and After \textsc{ITKI}}, each T2I model shows significant improvement.}
    \label{fig:training-free example}
\end{wrapfigure}

\begin{figure}
  \centering
\includegraphics[width=1\textwidth]{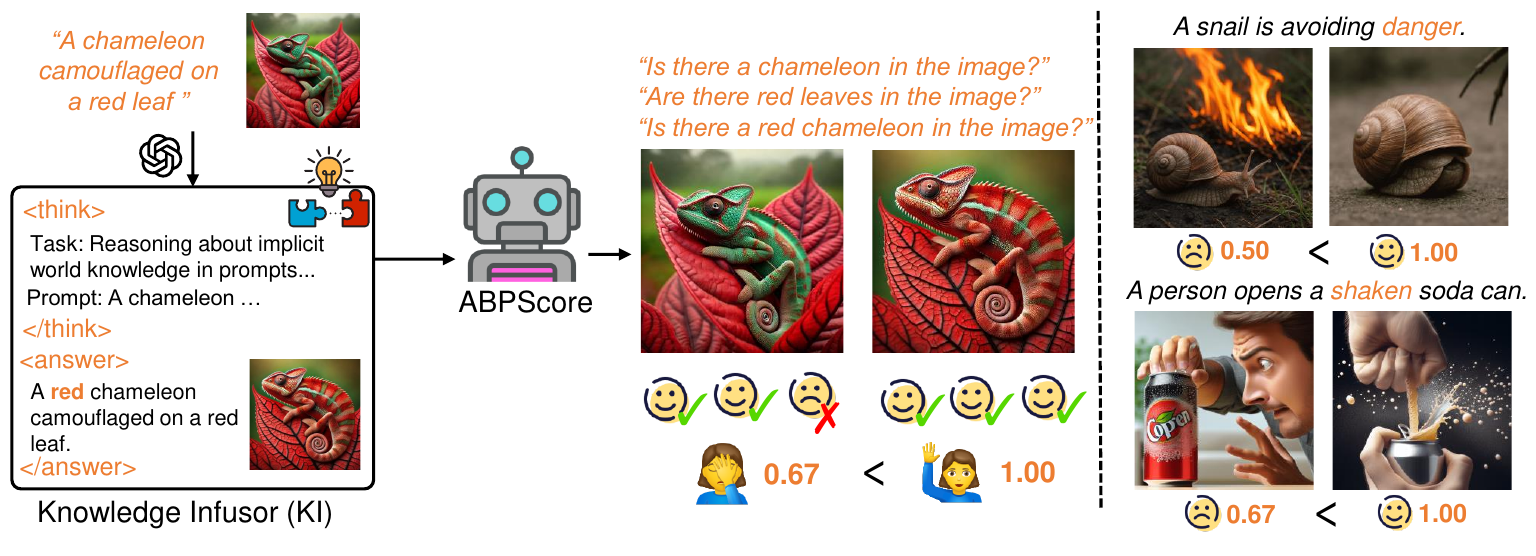}
  \caption{\textbf{Examples optimized through \textsc{ITKI}.} We utilize GPT-4o as a Knowledge Infusor (KI) to extract world knowledge beyond user-provided prompts. Examples optimized with \textsc{ITKI} achieve higher scores, and the corresponding images better represent the integration of world knowledge.}
  \label{figure: challenging example}
\end{figure}

After analyzing the results in Section \ref{sec:4.3}, we identified that T2I models exhibit limitations when generating images incorporating world knowledge beyond the provided prompts. 
However, it remains an open question whether explicitly describing the world knowledge beyond the provided prompts and reasoning for the expected visual outcomes can improve the \textsc{ABPScore} of existing T2I models. 
To address this, we propose an optimization strategy based on the inference-time scaling law \cite{ma2025inference}, called \textit{Inference-Time Knowledge Injection} (\textsc{ITKI}). This strategy does not require training the T2I model.
Specifically, we modify the pipeline in \textsc{ABP} to enhance prompts rather than propose the question, which is then named Knowledge Infusor (KI), into the T2I model to comprehend world knowledge into the given prompt. 
The detailed process is illustrated in Figure \ref{figure: challenging example}.
To validate the effectiveness of this strategy, we selected 200 challenging samples (with the lowest \textsc{ABPScore}) from the \textsc{ABP} and compared the generated images before and after adopting \textsc{ITKI}.
The experimental results are shown in Figure \ref{fig:training-free example}. By comparing the ABPScores before and after optimization, we observed a significant improvement of approximately 43\% across eight T2I models on average. 
This improvement can be attributed to the enhanced inference module's ability to better comprehend world knowledge, enabling the T2I models to generate images that align more closely with world knowledge, all without the need for additional training. In the Appendix, we also provide further experimental results conducted on additional prompts in \textsc{ABP}, along with results validated by existing benchmarks.

\section{Conclusion and Future Work}
In this work, we introduce \textsc{ABP}, a comprehensive benchmark designed to measure the alignment of generated images with real-world knowledge beyond the textual prompts.
\textsc{ABP} contains over 2,000 meticulously crafted prompts spanning six domains of world knowledge, along with an evaluation metric, \textsc{ABPScore}, which highly correlates with human judgments.
Through a comprehensive evaluation of eight popular T2I models using \textsc{ABP}, we find that even state-of-the-art models, such as GPT-4o, exhibit misalignment between the generated images and real-world knowledge beyond the textual prompts.
To address this issue, we introduce a training-free strategy, \textit{Inference-Time Knowledge Injection} (\textsc{ITKI}), to optimize 200 challenging samples in \textsc{ABP}. The results demonstrate an improvement of approximately 43\% in the \textsc{ABPScore} and notable improvements in existing benchmarks.
Through experimental analysis, we have identified that reasoning capabilities are crucial for developing T2I models that better align with world knowledge beyond the prompts provided by users. 
This insight provides a key direction for the future development of T2I models. This demonstrates the effectiveness of \textsc{ITKI} in enhancing the performance of T2I models without additional training.
We hope that \textsc{ABP} can contribute to the development and evaluation of more advanced T2I models, ultimately advancing the quality and reliability of visual generation. By providing a systematic and comprehensive approach to evaluating the alignment of generated images with world knowledge beyond the textual prompts, \textsc{ABP} offers a valuable resource for future research and the continuous improvement of T2I technology.

{
\small
\bibliographystyle{plainnat}  
\bibliography{main}  
}

\end{document}